\documentclass[runningheads]{llncs}
\usepackage{graphicx}
\usepackage{amsmath}
\usepackage{amsfonts}
\usepackage{booktabs}
\usepackage{subcaption}
\usepackage{xspace}
\usepackage{siunitx}
\usepackage{pgfplots}
\usepackage{tikz}
\usepackage{pstricks}
\usepackage[misc,geometry]{ifsym} 
\usepackage[capitalise]{cleveref}

\makeatletter
\DeclareRobustCommand\onedot{\futurelet\@let@token\@onedot}
\def\@onedot{\ifx\@let@token.\else.\null\fi\xspace}

\def\eg{\emph{e.g}\onedot} 
\def\ie{\emph{i.e}\onedot}

\def\etal{\emph{et al}\onedot}
\makeatother

\definecolor{oorange}{RGB}{255,102,0}

\newenvironment{customlegend}[1][]{%
	\begingroup
	\csname pgfplots@init@cleared@structures\endcsname
	\pgfplotsset{#1}%
}{%
	\csname pgfplots@createlegend\endcsname
\endgroup
}%
\def\addlegendimage{\csname pgfplots@addlegendimage\endcsname}

\usepackage{fancyhdr}
\fancypagestyle{specialfooter}{%
  \fancyhf{}
  
  \fancyfoot[L]{\scriptsize{$\copyright$ Pattern Recognition for Cultural Heritage (PatReCH) 2022 Workshop at the 26th International Conference on Pattern Recognition (ICPR) 2022, scheduled for August 21, 2022.}}
}
\begin{document}
\title{A Multi-modal Registration and Visualization Software Tool for Artworks using CraquelureNet}
\titlerunning{Registration and Visualization Software Tool for Artworks}
%
\author{Aline Sindel \and
Andreas Maier \and
Vincent Christlein}

\authorrunning{A. Sindel et al.}
%
\institute{Pattern Recognition Lab, FAU Erlangen-N\"urnberg, Germany\\
\email{aline.sindel@fau.de}
}
\maketitle              
%
\thispagestyle{specialfooter}
\begin{abstract}
For art investigations of paintings, multiple imaging technologies, such as visual light photography, infrared reflectography, ultraviolet fluorescence photography, and x-radiography are often used. For a pixel-wise comparison, the multi-modal images have to be registered. We present a registration and visualization software tool, that embeds a convolutional neural network to extract cross-modal features of the crack structures in historical paintings for automatic registration. The graphical user interface processes the user's input to configure the registration parameters and to interactively adapt the image views with the registered pair and image overlays, such as by individual or synchronized zoom or movements of the views. In the evaluation, we qualitatively and quantitatively show the effectiveness of our software tool in terms of registration performance and short inference time on multi-modal paintings and its transferability by applying our method to historical prints.

\keywords{Multi-modal registration  \and Visualization \and Convolutional neural networks.}
\end{abstract}
\section{Introduction}
In art investigations often multiple imaging systems, such as visual light photography (VIS), infrared reflectography (IRR), ultraviolet fluorescence photography (UV), and x-radiography (XR), are utilized. For instance, IR is used to reveal underdrawings, UV to visualize overpaintings and restorations, and XR to highlight white lead. 
Since the multi-modal images are acquired using different imaging systems, we have to take into account different image resolutions and varying viewpoints of the devices.
Thus, for a direct comparison on pixel level, image registration is crucial to align the multi-modal images.

\begin{figure}[t]
	\centering
\includegraphics[width=\textwidth]{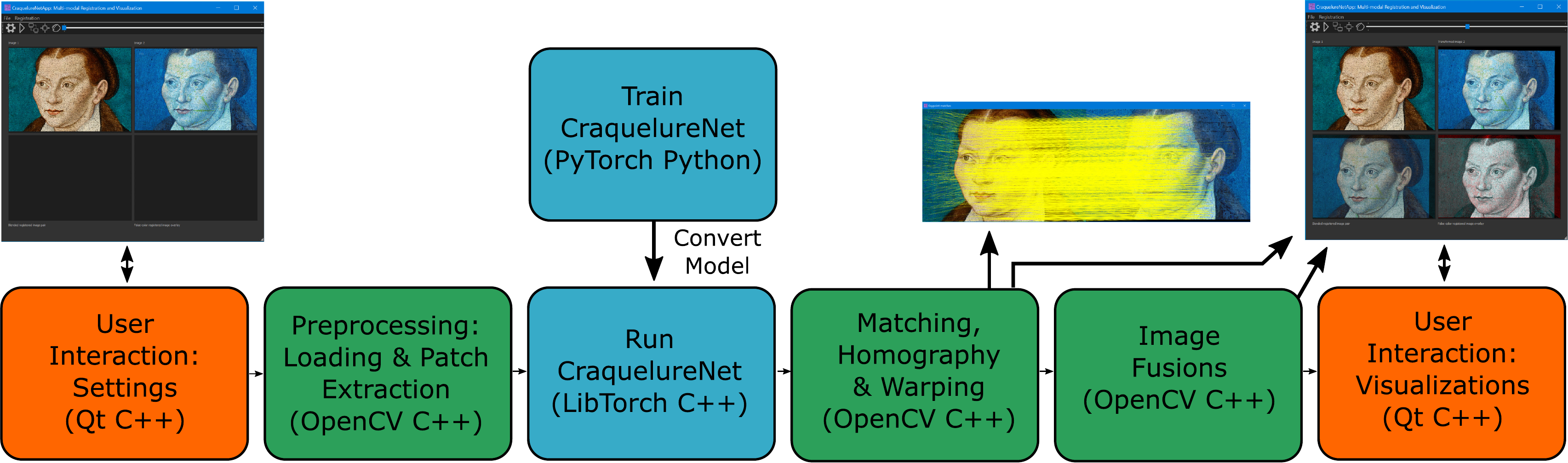}
	\caption{Our multi-modal registration and visualization tool CraquelureNetApp. Users can select the images and registration options in the GUI. The registration itself is performed fully automatically using the CraquelureNet, a CNN for multi-modal keypoint detection and description, which was ported from PyTorch to C++. Due to the high-resolution images of paintings, the CraquelureNet is applied patch-wise. The registration results are depicted inside the GUI and the users can interact with the visualizations, \eg joint zooming of the views. 
Image sources: Workshop Lucas Cranach the Elder, Katharina of Bora (Detail), visual light and UV-fluorescence photographs, captured by Ulrike Hügle, Stiftung Deutsches Historisches Museum, Berlin, Cranach Digital Archive, KKL-No IV.M21b, all rights reserved
}
\label{fig-01:overview}
\end{figure} 

Image registration methods for art imaging can mainly be split into intensity\mbox{-,} control point- and feature-based approaches.
The intensity-based method of Cappellini~\etal~\cite{CappelliniV2005} uses mutual information to iteratively register multispectral images.
The control point- and feature-based methods compute a geometric transform based on detected correspondences in both images.
Murashov~\cite{MurashovD2011} detects local grayscale maxima as control points in VIS-XR image pairs and applies coherent point drift~\cite{MyronenkoA2010}.
Conover \etal~\cite{ConoverDM2015} uses  Wavelet transform, phase correlation, and disparity filtering for control point detection in sub-images of VIS, IR, and XR.
The feature-based method of Zacharopoulos~\etal~\cite{ZacharopoulosA2018} uses SIFT~\cite{LoweDG2004} to align multispectral images of artworks.
CraquelureNet~\cite{SindelA2021} uses a convolutional neural network (CNN) to detect keypoints and descriptors based on branching points in the crack structure (craquelure) of the paint, as it is visible by all imaging systems, in contrast to the depicted image content that can be very different in the multi-modal images. 

More and more museums or art projects provide the ability to inspect their artworks in interactive website viewers. Specifically for multi-modal images website viewer have been designed that allow a synchronized scrolling and zooming of the image views~\cite{BoschProject,FransenB2020} or a curtain viewer~\cite{BoschProject} that allows to interactively inspect multiple images in a single view. For these projects, the specific multi-modal images of artworks were pre-registered offline.

For the daily work, it would be practical for art technologists and art historians to have a tool, with which they can easily perform the registration themselves and also can interactively inspect the registered images.
In the field of medical 2D-2D or 3D-3D registration, there are open source tools such as MITK~\cite{SteinD2010} that provides a graphical user interface (GUI) application for iterative rigid and affine registration and also a developer software framework. Since these software tools are very complex, domain knowledge is required to adapt the algorithms for multi-modal registration of paintings.
The ImageOverlayApp~\cite{SindelA2020} is a small and easy to handle GUI application that allows the direct comparison of two registered artworks using image superimposition and blending techniques. However, here as well as in the online viewers, image registration is not provided.

In this paper, we propose a software tool for automatic multi-modal image registration and visualization of artworks. We designed a GUI to receive the user's registration settings and to provide an interactive display to show the registration results, such as superimposition and blending of the registered image pair, and both, individual and synchronized zooming and movement of the image views. As registration method, we integrate the keypoint detection and description network CraquelureNet into our application by porting it to C++. A quantitative evaluation and qualitative examples show the effective application of our software tool on our multi-modal paintings dataset and its transferability to our historical prints dataset.

\section{Methods}
In this section, the software tool for multi-modal registration and visualization is described.

\subsection{Overview of the Registration and Visualization Tool}
In Fig.~\ref{fig-01:overview}, the main building blocks of the CraquelureNetApp are shown, consisting of the GUI, the preprocessing such as data loading and patch extraction, the actual registration method, which is split into the keypoint detection and description network CraquelureNet that was ported from PyTorch to LibTorch and into descriptor matching, homography estimation, and image warping, and the computation of visualizations of the registration results and its interactions with the user.
 
\subsection{Registration Method}
CraquelureNet~\cite{SindelA2021} is composed of a ResNet backbone and a keypoint detection and a keypoint description head. The network is trained on small $32 \times 32 \times 3$ sized image patches using a multi-task loss for both heads. The keypoint detection head is optimized using binary cross-entropy loss to classify the small patches into ``craquelure'' and ``background'', i.e. whether the center of the patch (a) contains a branching or sharp bend of a crack structure (``craquelure'') or (b) contains it only in the periphery or not at all (``background''). The keypoint description head is trained using bidirectional quadruplet loss~\cite{SindelA2021} to learn cross-modal descriptors. Using online hard negative mining, for a positive keypoint pair the hardest non-matching descriptors are selected in both directions within the batch. Then, the positive distances (matching keypoint pairs) are minimized while the negative distances (hardest non-matching keypoint pairs) are maximized.

For inference, larger image input sizes can be fed to the fully convolutional network. Due to the architectural design, the prediction of the keypoint detection head and description head are of lower resolution than the input image and hence are upscaled by a factor of $4$ using bicubic interpolation for the keypoint heatmap and bilinear interpolation for the descriptors based on the extracted keypoint positions. Keypoints are extracted using non-maximum suppression and by selecting all keypoints with a confidence score higher than $\tau_\text{kp}$ based on the keypoint heatmap. 
Since the images can be very large, CraquelureNet is applied patch-based and the keypoints of all patches are merged and reduced to the $N_\text{max}$ keypoints with the highest confidence score. Then, mutual nearest neighbor descriptor matching is applied and Random Sample Consensus (RANSAC)~\cite{FischlerMA1981} is used for homography estimation~\cite{SindelA2021}.

\begin{figure}[t]
	\centering
\includegraphics[width=\textwidth]{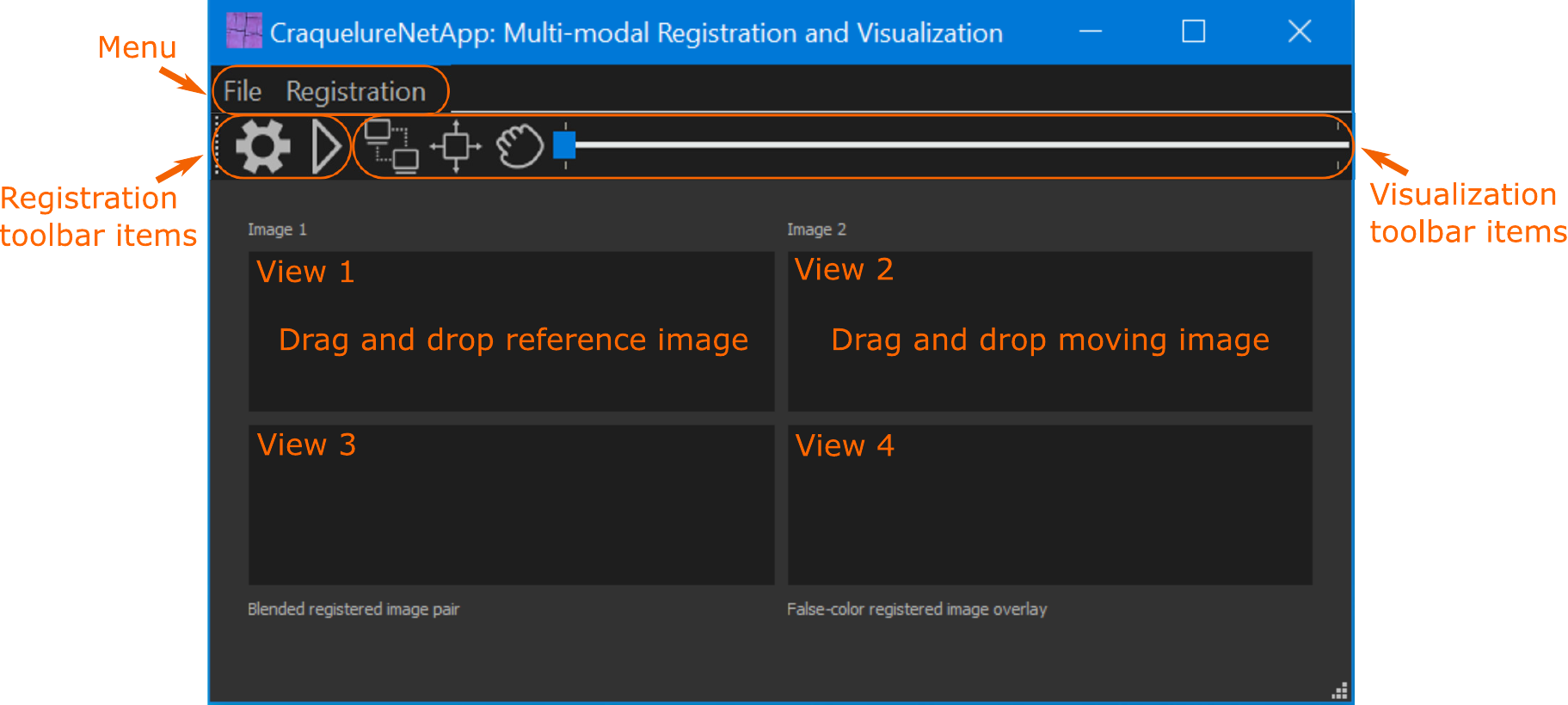}
	\caption{The graphical user interface (GUI) of our CraquelureNetApp. The different elements of the GUI are marked in orange: the menu, the toolbar items for registration and visualization and the views.}
	\label{fig-02:gui}
\end{figure} 

CraquelureNet~\cite{SindelA2021} is completely implemented in Python using PyTorch and OpenCV. To transfer the trained PyTorch model to C++, we made use of the TorchScript conversion functionality and restructured the source code accordingly using a combination of TorchScript modules and Tracing. We have directly reimplemented the other parts of the registration pipeline in C++ using the same OpenCV functions as in Python.

For homography estimation, we provide the option to choose between different robust estimators of the OpenCV library: RANSAC~\cite{FischlerMA1981} which is already used in~\cite{SindelA2021,SindelA2022} is the default option. RANSAC is an iterative method to estimate a model, here homography, for data points that contain outliers. In each iteration, a model is estimated on a random subset of the data and is scored using all data points, \ie the number of inliers for that model is computed. In the end, the best model is selected based on the largest inlier count and can optionally be refined.  
Universal RANSAC (USAC)~\cite{RaguramR2013} includes different RANSAC variants into a single framework, \eg different sampling strategies, or extend the verification step by also checking for degeneracy. We use the USAC default method of OpenCV 4.5.5 that applies iterative local optimization to the so far best models~\cite{ChumO2003}.
Further, we enable two RANSAC variants which are also included in the OpenCV USAC framework:
Graph-Cut RANSAC (GC-RANSAC)~\cite{BarathD2018} uses the graph-cut algorithm for the local optimization and MAGSAC++~\cite{BarathD2020} applies a scoring function that does not require a threshold and a novel marginalization method.

\begin{figure}[t]
	\centering
\includegraphics[width=\textwidth]{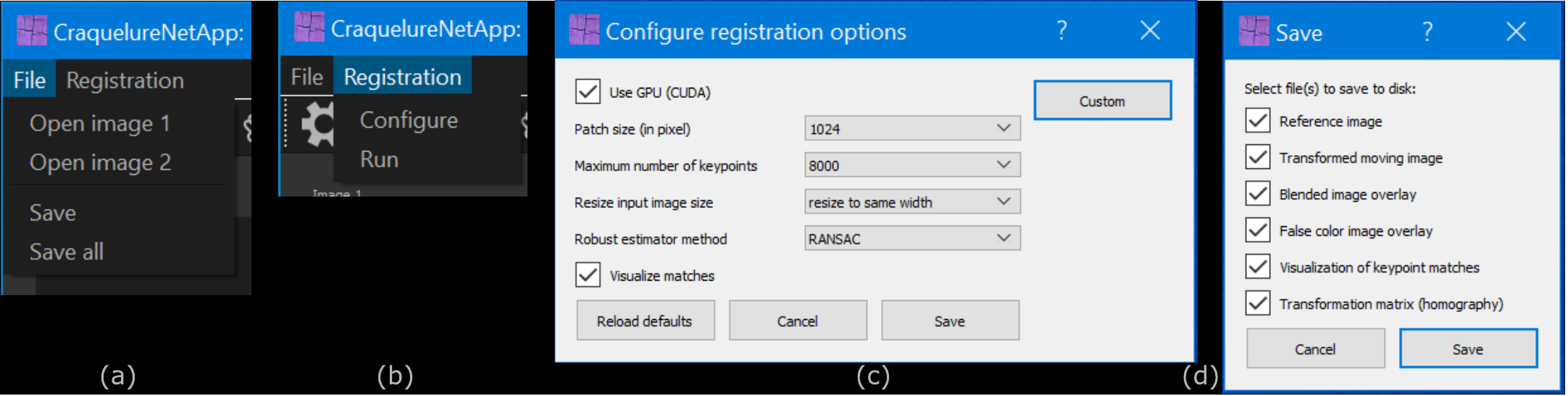}
	\caption{The menu and dialogs to configure registration and save options.}
	\label{fig-03:menu_dialogs}
\end{figure} 

\subsection{Graphical User Interface}
The GUI is implemented in C++ as a Qt5 desktop application. In Fig.~\ref{fig-02:gui} the main GUI elements are marked in orange. The user can interact with the GUI using the menu or the toolbar that provides items specified for the registration and visualization task or can also directly interact with the views. The size of the views depends on the size of the application window.

\subsubsection{User interaction prior to registration:} To perform registration, two images have to be selected. They can either be chosen via the menu (Fig.~\ref{fig-03:menu_dialogs}a) using a file opener or per drag and drop of the image file to the image view. The selected images are visualized in the two image views in the top row. 
Per click on the configure button in the menu (Fig.~\ref{fig-03:menu_dialogs}b) or on the registration toolbar (``gear'' icon), a user dialog (Fig.~\ref{fig-03:menu_dialogs}c) is opened to choose the registration options, such as patch size, number of maximum keypoints $N_\text{max}$, image input size, method for homography estimation (RANSAC, USAC, GC-RANSAC, or MAGSAC++), whether to run on the GPU (using CUDA) or on the CPU and whether to visualize keypoint matches.
Custom settings of patch size, input image size, and maximum number of keypoints are also possible when the predefined selections are not sufficient. The settings are saved in a configuration file to remember the user's choice. It is always possible to restore the default settings. The registration is started by clicking the run button in the menu (Fig.~\ref{fig-03:menu_dialogs}b) or in the toolbar (``triangle'' icon).

\subsubsection{Visualizations and user interaction:} We include two different superimposition techniques to visually compare the registration results, a false-color image overlay (red-cyan) of the registered pair and a blended image overlay, similarly to~\cite{SindelA2020}. After registration, the views in the GUI are updated with the transformed moving image (view 2) and the image fusions in the bottom row (view 3 and 4). The red-cyan overlay (view 4), stacks the reference image (view 1) into the red and the transformed moving image (view 2) into the green and blue channel. The blended image (view 3) is initially shown with alpha value 0.5. By moving the slider in the visualization toolbar, the user can interactively blend between both registered images. For the optional visualization of the keypoint matches, a separate window is opened that shows both original images with superimposed keypoints as small blue circles and matches as yellow lines using OpenCV.

Additionally to the image overlays, we have implemented a synchronization feature of the views that enriches the comparison. It can be activated by the ``connect views'' icon in the visualization toolbar. All interactions with one view are propagated to the other views. Using the mouse wheel the view is zoomed in or out with the focus at the current mouse position. Using the arrow keys the image view can be shifted in all directions. By activating the ``hand mouse drag'' item in the toolbar, the view can be shifted around arbitrarily. By pressing the ``maximize'' icon in the toolbar, the complete image is fitted into the view area preserving aspect ratio. This can also be useful after asynchronous movement of single views to reset them to a common basis. 

To save the registration results to disk, the user can click the ``save'' button in the menu (Fig.~\ref{fig-03:menu_dialogs}a) to choose the files to save in a dialog (Fig.~\ref{fig-03:menu_dialogs}d) or the ``save all'' button to save them all. 

\begin{figure}[t]
	\centering
\includegraphics[width=\textwidth]{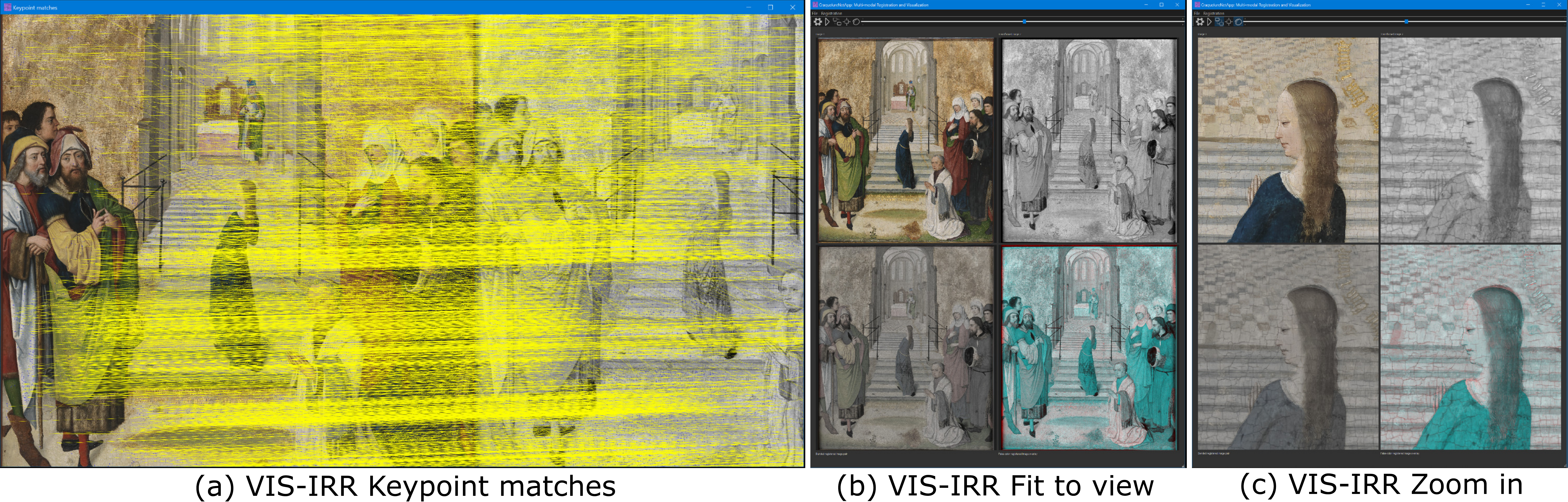}
	\caption{VIS-IRR registration of paintings using CraquelureNetApp. In (a) the keypoint matches between the reference image (VIS) and moving image (IRR) are visualized as yellow lines. In (b) the registered image pair and the blended overlay and the false color overlay are depicted as complete images and in (c) they are synchronously zoomed in.
	Image sources: Meister des Marienlebens, Tempelgang Mari\"a, visual light photograph and infrared reflectogram, Germanisches Nationalmuseum, Nuremberg, on loan from Wittelsbacher Ausgleichsfonds/Bayerische Staats\-gem\"alde\-samm\-lungen, \mbox{Gm 19}, all rights reserved}
	\label{fig-04:vis_irr}
\end{figure} 

\begin{figure}[t]
	\centering
\includegraphics[width=\textwidth]{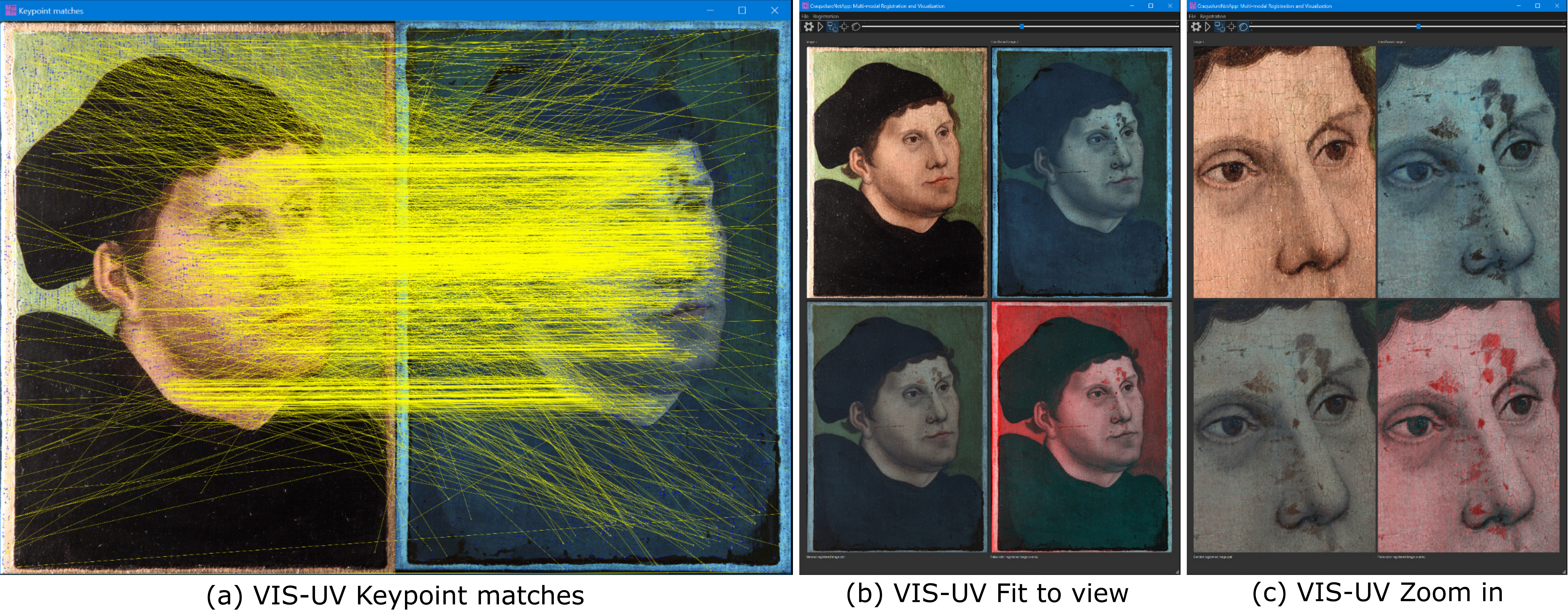}
	\caption{VIS-UV registration of paintings using CraquelureNetApp. In (a) the keypoint matches of VIS and UV are densely concentrated at the  craquelure in the facial region. (b) and (c) show the registered pair and image fusions of the complete images and synchronized details.	
	Image sources: Workshop Lucas Cranach the Elder or Circle, Martin Luther, visual light photograph, captured by Gunnar Heydenreich and UV-fluorescence photograph, captured by Wibke Ottweiler, Lutherhaus Wittenberg, Cranach Digital Archive, KKL-No I.6M3, all rights reserved} 
	\label{fig-05:vis_uv}
\end{figure} 

\begin{figure}[t]
	\centering
\includegraphics[width=\textwidth]{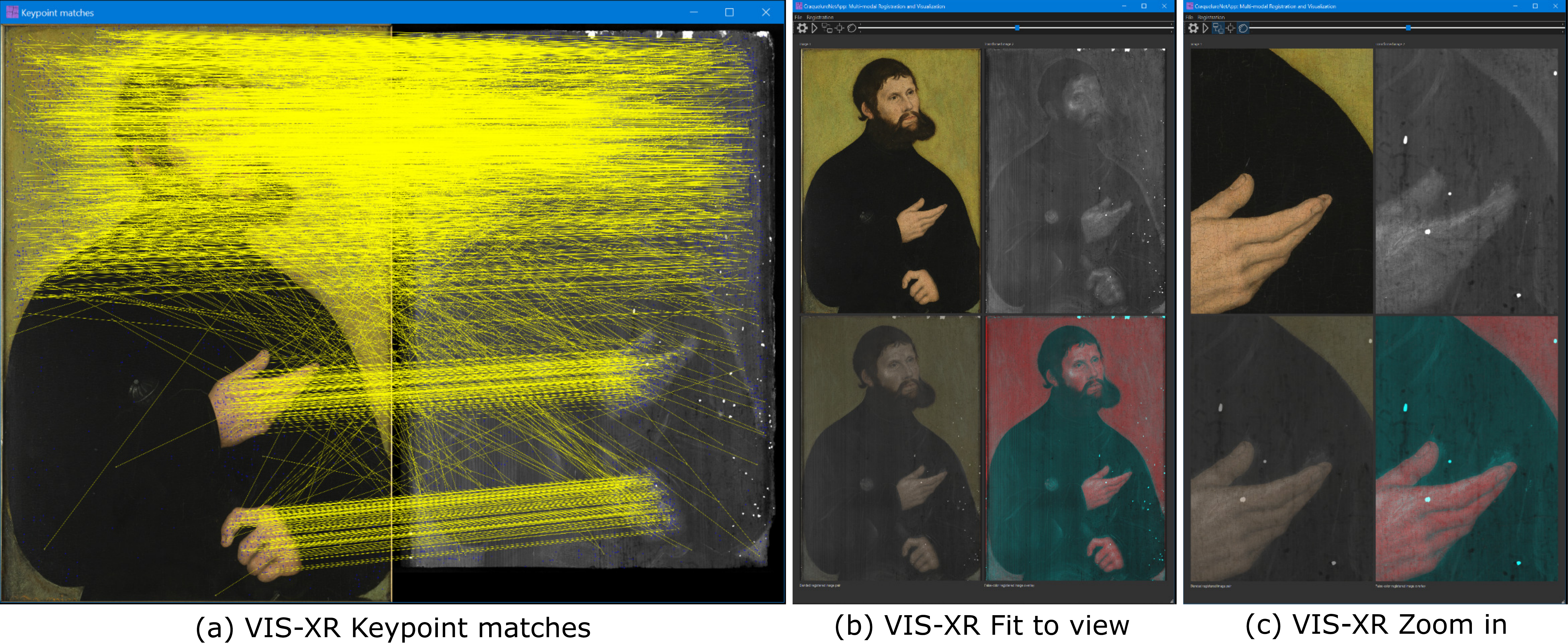}
	\caption{VIS-XR registration of paintings using CraquelureNetApp. In (a) the keypoint matches of VIS and XR are mostly concentrated at the craquelure in the lighter image area such as the face, hands, and green background. (b) and (c) show the qualitative registrations results as overall view and as synchronized zoomed details.
	Image sources: Lucas Cranach the Elder, Martin Luther as ``Junker Jörg'', visual light photograph Klassik Stiftung Weimar, Museum, X-radiograph HfBK Dresden (Mohrmann, Riße), Cranach Digital Archive, KKL-No II.M2, all rights reserved}
	\label{fig-06:vis_xr}
\end{figure} 

\begin{table}[t]
\centering
\caption{Quantitative evaluation for multi-modal registration of paintings for the VIS-IRR, VIS-UV, and VIS-XR test datasets (each 13 image pairs) using CraquelureNet with different robust homography estimation methods. Registration results are evaluated with the success rates (SR) of mean Euclidean error (ME) and maximum Euclidean error (MAE) of the control points for different error thresholds $\epsilon$. Best results are highlighted in bold.} 
\label{tab-01}
\scriptsize
\begin{tabular*}{\textwidth}{l @{\extracolsep{\fill}} ll rrr rrr}
\toprule
Multi-modal & CraquelureNet & Homography  & \multicolumn{3}{c}{SR of ME [\%] $\uparrow$} & \multicolumn{3}{c}{SR of MAE [\%] $\uparrow$}\\
Dataset & Model & Method & $\epsilon=3$ & $\epsilon=5 $ & $\epsilon=7$ & $\epsilon=6$ & $\epsilon=8$ & $\epsilon=10$ \\ 
\midrule
 & PyTorch (Python) & RANSAC & 84.6 & \textbf{100.0} & \textbf{100.0} & 38.5 & \textbf{69.2} & \textbf{84.6} \\ 
 & LibTorch (C++) & RANSAC & \textbf{92.3} & 92.3 & \textbf{100.0} & 38.5 & \textbf{69.2} & \textbf{84.6} \\ 
VIS-IRR & LibTorch (C++) & USAC & \textbf{92.3} & 92.3 & 92.3 & \textbf{46.2} & \textbf{69.2} & \textbf{84.6} \\ 
 & LibTorch (C++) & GC-RANSAC & \textbf{92.3} & 92.3 & 92.3 & \textbf{46.2} & \textbf{69.2} & \textbf{84.6} \\ 
 & LibTorch (C++) & MAGSAC++ & \textbf{92.3} & 92.3 & 92.3 & \textbf{46.2} & \textbf{69.2} & \textbf{84.6} \\ 
\midrule
& PyTorch (Python) & RANSAC  & \textbf{92.3} & \textbf{100.0} & \textbf{100.0} & 46.2 & 53.8 & 61.5 \\ 
 & LibTorch (C++) & RANSAC & 84.6 & 92.3 & \textbf{100.0} & \textbf{53.8} & 53.8 & 53.8 \\ 
VIS-UV & LibTorch (C++) & USAC & \textbf{92.3} & 92.3 & 92.3 & \textbf{53.8} & \textbf{69.2} & \textbf{69.2} \\ 
 & LibTorch (C++) & GC-RANSAC & \textbf{92.3} & 92.3 & 92.3 & \textbf{53.8} & \textbf{69.2} & \textbf{69.2} \\ 
 & LibTorch (C++) & MAGSAC++ & 84.6 & 92.3 & 92.3 & 46.2 & 61.5 & \textbf{69.2} \\ 
\midrule
 & PyTorch (Python) & RANSAC  & 69.2 & \textbf{84.6} & 84.6 & 23.1 & 38.5 & 61.5 \\ 
 & LibTorch (C++) & RANSAC & 53.8 & \textbf{84.6} & \textbf{92.3} & \textbf{30.8} & 46.2 & 61.5 \\ 
VIS-XR & LibTorch (C++) & USAC & 76.9 & \textbf{84.6} & 84.6 & 23.1 & 38.5 & \textbf{76.9} \\ 
 & LibTorch (C++) & GC-RANSAC & 76.9 & \textbf{84.6} & 84.6 & 23.1 & 38.5 & \textbf{76.9} \\ 
 & LibTorch (C++) & MAGSAC++ & \textbf{84.6} & \textbf{84.6} & 84.6 & 15.4 & \textbf{53.8} & \textbf{76.9} \\ 
\bottomrule 
\end{tabular*}
\end{table}

\begin{figure}[t]								 
\centering	
\begin{tikzpicture}
	\begin{customlegend}[
			legend entries={SIFT,D2-Net,SuperPoint,CraquelureNet (C++)},
			legend style={
				/tikz/every even column/.append	style={column sep=.75cm}},
			legend columns=-1,
			legend cell align=left]
		\addlegendimage{oorange,mark=square*}
		\addlegendimage{red,mark=square*}		
		\addlegendimage{blue,mark=square*}
		\addlegendimage{cyan,mark=square*}				
	\end{customlegend}
\end{tikzpicture}

\subcaptionbox{VIS-IRR\label{fig-07:paint_VIS-IRR_me}}{
\begin{tikzpicture}
    \begin{axis}[
        width  = 0.33\textwidth,
        height = 3cm,
        major x tick style = transparent,
        ybar,
        bar width=0.15cm,
    		x=1.3cm, 
    		enlarge x limits={abs=0.7cm},
    		enlarge y limits=false,
        ymajorgrids = true,
        ylabel = {SR of ME [\%]},
      	ymin=0,
        ymax=100,
        symbolic x coords= {$\epsilon=3$,$\epsilon=7$},
        xtick = data,
        nodes near coords={\pgfmathprintnumber[fixed zerofill,precision=2]\pgfplotspointmeta},
        every node near coord/.append style={font=\tiny, xshift = 0 , rotate=90, anchor=west},
        nodes near coords align = {center},
        ]
        \addplot [style={oorange,fill=oorange,mark=none}]
        table[x=Metric, y=SIFT] 
        {tables/paintings_VIS-IRR_success_rates_me.dat};             
        \addplot [style={red,fill=red,mark=none}]
        table[x=Metric, y=D2-Net] 
        {tables/paintings_VIS-IRR_success_rates_me.dat};  
        \addplot [style={blue,fill=blue,mark=none}]
        table[x=Metric, y=SuperPoint] 
        {tables/paintings_VIS-IRR_success_rates_me.dat}; 
        \addplot [style={cyan,fill=cyan,mark=none}]
        table[x=Metric, y=CraquelureNetC++] 
        {tables/paintings_VIS-IRR_success_rates_me.dat};                      
	\end{axis}
\end{tikzpicture}	
	}
\subcaptionbox{VIS-UV\label{fig-07:paint_VIS-UV_me}}{
\begin{tikzpicture}
    \begin{axis}[
        width  = 0.33\textwidth,
        height = 3cm,
        major x tick style = transparent,
        ybar,
        bar width=0.15cm,
    		x=1.3cm, 
    		enlarge x limits={abs=0.7cm},
    		enlarge y limits=false,
        ymajorgrids = true,
      	ymin=0,
        ymax=100,
        symbolic x coords= {$\epsilon=3$,$\epsilon=7$},
        xtick = data,
        nodes near coords={\pgfmathprintnumber[fixed zerofill,precision=2]\pgfplotspointmeta},
        every node near coord/.append style={font=\tiny, xshift = 0 , rotate=90, anchor=west},
        nodes near coords align = {center},
        ]
        \addplot [style={oorange,fill=oorange,mark=none}]
        table[x=Metric, y=SIFT] 
        {tables/paintings_VIS-UV_success_rates_me.dat};             
        \addplot [style={red,fill=red,mark=none}]
        table[x=Metric, y=D2-Net] 
        {tables/paintings_VIS-UV_success_rates_me.dat};  
        \addplot [style={blue,fill=blue,mark=none}]
        table[x=Metric, y=SuperPoint] 
        {tables/paintings_VIS-UV_success_rates_me.dat}; 
        \addplot [style={cyan,fill=cyan,mark=none}]
        table[x=Metric, y=CraquelureNetC++] 
        {tables/paintings_VIS-UV_success_rates_me.dat};                      
	\end{axis}
\end{tikzpicture}	
	}
\subcaptionbox{VIS-XR\label{fig-07:paint_VIS-XR_me}}{
\begin{tikzpicture}
    \begin{axis}[
        width  = 0.33\textwidth,
        height = 3cm,
        major x tick style = transparent,
        ybar,
        bar width=0.15cm,
    		x=1.3cm, 
    		enlarge x limits={abs=0.7cm},
    		enlarge y limits=false,
        ymajorgrids = true,
      	ymin=0,
        ymax=100,
        symbolic x coords= {$\epsilon=3$,$\epsilon=7$},
        xtick = data,
        nodes near coords={\pgfmathprintnumber[fixed zerofill,precision=2]\pgfplotspointmeta},
        every node near coord/.append style={font=\tiny, xshift = 0 , rotate=90, anchor=west},
        nodes near coords align = {center},
        ]
        \addplot [style={oorange,fill=oorange,mark=none}]
        table[x=Metric, y=SIFT] 
        {tables/paintings_VIS-XR_success_rates_me.dat};             
        \addplot [style={red,fill=red,mark=none}]
        table[x=Metric, y=D2-Net] 
        {tables/paintings_VIS-XR_success_rates_me.dat};  
        \addplot [style={blue,fill=blue,mark=none}]
        table[x=Metric, y=SuperPoint] 
        {tables/paintings_VIS-XR_success_rates_me.dat}; 
        \addplot [style={cyan,fill=cyan,mark=none}]
        table[x=Metric, y=CraquelureNetC++] 
        {tables/paintings_VIS-XR_success_rates_me.dat};                      
	\end{axis}
\end{tikzpicture}		
	}	
	
\subcaptionbox{VIS-IRR\label{fig-07:paint_VIS-IRR_mae}}{
\begin{tikzpicture}
    \begin{axis}[
        width  = 0.33\textwidth,
        height = 3cm,
        major x tick style = transparent,
        ybar,
        bar width=0.15cm,
    		x=1.3cm, 
    		enlarge x limits={abs=0.7cm},
    		enlarge y limits=false,
        ymajorgrids = true,
      	ymin=0,
        ymax=100,
        ylabel = {SR of MAE [\%]},
        symbolic x coords= {$\epsilon=6$,$\epsilon=10$},
        xtick = data,
        nodes near coords={\pgfmathprintnumber[fixed zerofill,precision=2]\pgfplotspointmeta},
        every node near coord/.append style={font=\tiny, xshift = 0 , rotate=90, anchor=west},
        nodes near coords align = {center},
        ]
        \addplot [style={oorange,fill=oorange,mark=none}]
        table[x=Metric, y=SIFT] 
        {tables/paintings_VIS-IRR_success_rates_mae.dat};             
        \addplot [style={red,fill=red,mark=none}]
        table[x=Metric, y=D2-Net] 
        {tables/paintings_VIS-IRR_success_rates_mae.dat};  
        \addplot [style={blue,fill=blue,mark=none}]
        table[x=Metric, y=SuperPoint] 
        {tables/paintings_VIS-IRR_success_rates_mae.dat}; 
        \addplot [style={cyan,fill=cyan,mark=none}]
        table[x=Metric, y=CraquelureNetC++] 
        {tables/paintings_VIS-IRR_success_rates_mae.dat};                      
	\end{axis}
\end{tikzpicture}	
	}
\subcaptionbox{VIS-UV\label{fig-07:paint_VIS-UV_mae}}{
\begin{tikzpicture}
    \begin{axis}[
        width  = 0.33\textwidth,
        height = 3cm,
        major x tick style = transparent,
        ybar,
        bar width=0.15cm,
    		x=1.3cm, 
    		enlarge x limits={abs=0.7cm},
    		enlarge y limits=false,
        ymajorgrids = true,
      	ymin=0,
        ymax=100,
        symbolic x coords= {$\epsilon=6$,$\epsilon=10$},
        xtick = data,
        nodes near coords={\pgfmathprintnumber[fixed zerofill,precision=2]\pgfplotspointmeta},
        every node near coord/.append style={font=\tiny, xshift = 0 , rotate=90, anchor=west},
        nodes near coords align = {center},
        ]
        \addplot [style={oorange,fill=oorange,mark=none}]
        table[x=Metric, y=SIFT] 
        {tables/paintings_VIS-UV_success_rates_mae.dat};             
        \addplot [style={red,fill=red,mark=none}]
        table[x=Metric, y=D2-Net] 
        {tables/paintings_VIS-UV_success_rates_mae.dat};  
        \addplot [style={blue,fill=blue,mark=none}]
        table[x=Metric, y=SuperPoint] 
        {tables/paintings_VIS-UV_success_rates_mae.dat}; 
        \addplot [style={cyan,fill=cyan,mark=none}]
        table[x=Metric, y=CraquelureNetC++] 
        {tables/paintings_VIS-UV_success_rates_mae.dat};                      
	\end{axis}
\end{tikzpicture}	
	}
\subcaptionbox{VIS-XR\label{fig-07:paint_VIS-XR_mae}}{
\begin{tikzpicture}
    \begin{axis}[
        width  = 0.33\textwidth,
        height = 3cm,
        major x tick style = transparent,
        ybar,
        bar width=0.15cm,
    		x=1.3cm, 
    		enlarge x limits={abs=0.7cm},
    		enlarge y limits=false,
        ymajorgrids = true,
      	ymin=0,
        ymax=100,
        symbolic x coords= {$\epsilon=6$,$\epsilon=10$},
        xtick = data,
        nodes near coords={\pgfmathprintnumber[fixed zerofill,precision=2]\pgfplotspointmeta},
        every node near coord/.append style={font=\tiny, xshift = 0 , rotate=90, anchor=west},
        nodes near coords align = {center},
        ]
        \addplot [style={oorange,fill=oorange,mark=none}]
        table[x=Metric, y=SIFT] 
        {tables/paintings_VIS-XR_success_rates_mae.dat};             
        \addplot [style={red,fill=red,mark=none}]
        table[x=Metric, y=D2-Net] 
        {tables/paintings_VIS-XR_success_rates_mae.dat};  
        \addplot [style={blue,fill=blue,mark=none}]
        table[x=Metric, y=SuperPoint] 
        {tables/paintings_VIS-XR_success_rates_mae.dat}; 
        \addplot [style={cyan,fill=cyan,mark=none}]
        table[x=Metric, y=CraquelureNetC++] 
        {tables/paintings_VIS-XR_success_rates_mae.dat};                      
	\end{axis}
\end{tikzpicture}		
	}
\caption{Quantitative evaluation for our multi-modal paintings dataset using SR of ME with $\epsilon=3,7$ and SR of MAE with $\epsilon=6,10$.}
\label{fig-07:eval_paintings_sota}
\end{figure}
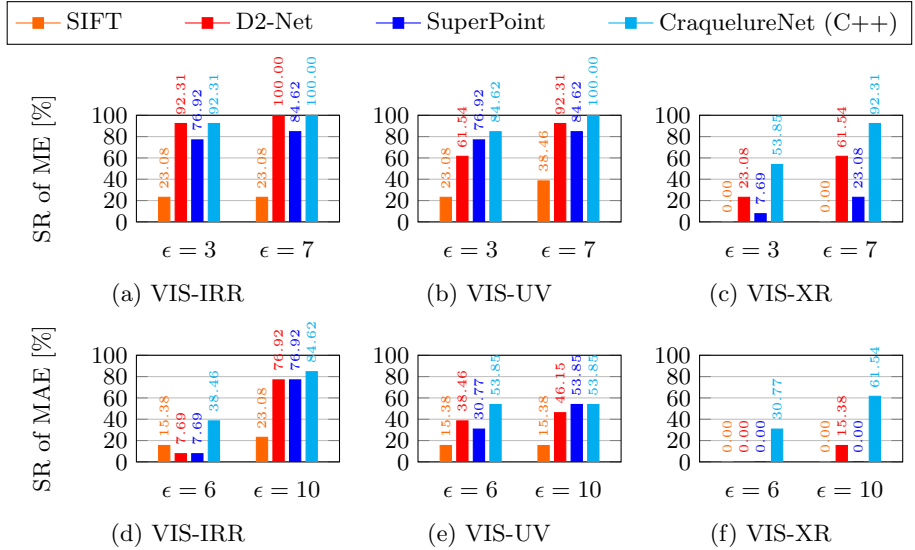 

\begin{figure}[t]
	\centering
\includegraphics[width=\textwidth]{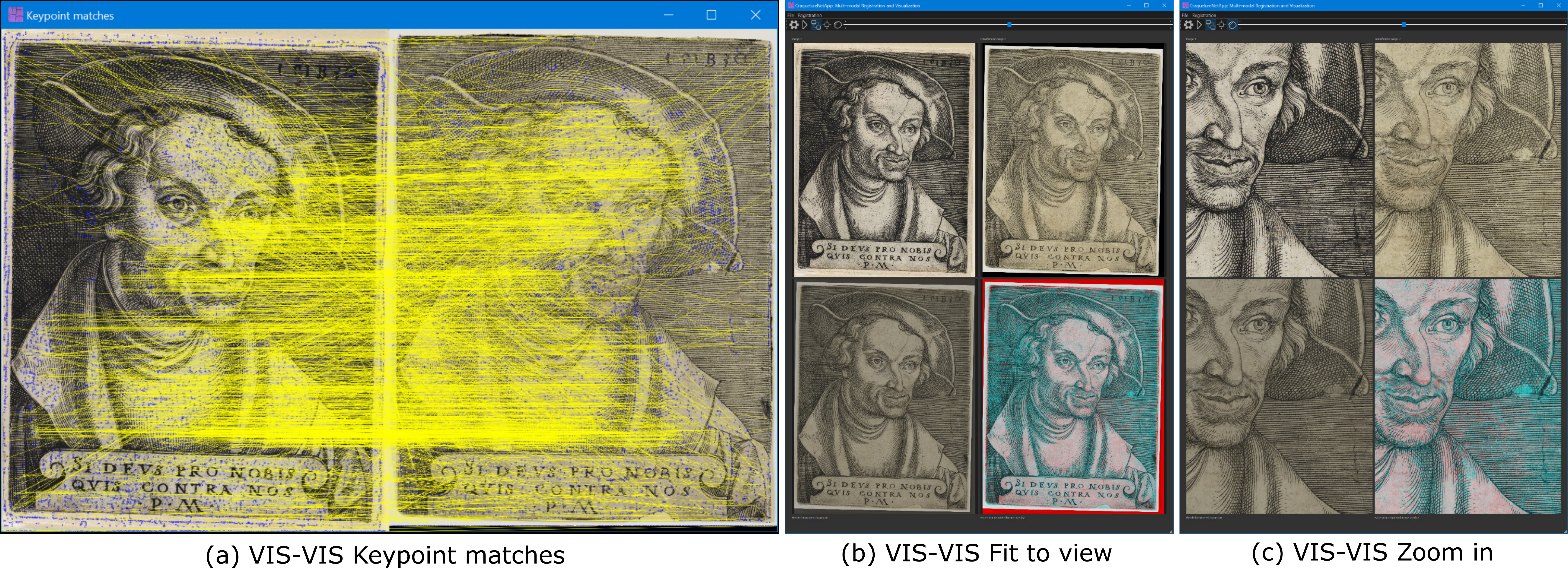}
	\caption{VIS-VIS registration of prints using CraquelureNetApp. CraquelureNet detects a high number of good matches for the two prints in (a), although it did not see images of prints during training. (b) and (c) show the qualitative registration results as overall and detail views.
	Image sources: Monogramist IB (Georg Pencz), Philipp Melanchthon with beret and cloak, (left) Germanisches Nationalmuseum, Nuremberg, K 21148, captured by Thomas Klinke and (right) Klassik Stiftung Weimar, Museen, Cranach Digital Archive, DE{\_}KSW{\_}Gr-2008-1858, all rights reserved}
	\label{fig-08:print}
\end{figure}

\section{Applications and Evaluation}
In this section, we apply the CraquelureNetApp to multi-modal images of paintings and to images of different prints of the same motif. We use the test parameters as defined in~\cite{SindelA2021}: GPU, patch size of $1024$, $N_\text{max}=8000$, $\tau_\text{kp}=0$, resize to same width, and RANSAC with a reprojection error threshold $\tau_\text{reproj}=5$.

For the quantitative evaluation, we measure the registration performance based on the success rate of successfully registered images. The success rate is computed by calculating the percentage of image pairs for which the error distance of manual labeled control points of the registered pair using the predicted homography is less or equal to an error threshold $\epsilon$.
As metric for the error distance, we use the mean Euclidean error (ME) and maximum Euclidean error (MAE)~\cite{SindelA2022}.

As comparison methods, we use the conventional keypoint and feature descriptor SIFT~\cite{LoweDG2004} and the pretrained models of the two deep learning methods, SuperPoint~\cite{DeToneD2018} and D2-Net~\cite{DusmanuM2019}, which we apply patch-based to both the paintings and prints. SuperPoint is a CNN with a keypoint detection head and a keypoint description head and is trained in a self-supervised manner by using homographic warpings. D2-Net simultaneously learns keypoint detection and description with one feature extraction CNN, where keypoints and descriptors are extracted from the same set of feature maps.
For all methods, we use the same test settings as for our method (patch size, $N_\text{max}$, RANSAC). 

\subsection{Multi-modal Registration of Historical Paintings}
The pretrained CraquelureNet~\cite{SindelA2021}, which we embedded into our registration tool, was trained using small patches extracted from multi-modal images of 16th century portraits by the workshop of Lucas Cranach the Elder and large German panel paintings from the 15th to 16th century. We use the images of the test split from these multi-modal datasets (13 pairs per domain: VIS-IRR, VIS-UV, VIS-XR) and the corresponding manually labeled control point pairs (40 point pairs per image pair)~\cite{SindelA2021} to test the CraquelureNetApp. 

The qualitative results of three examples (VIS-IRR, VIS-UV and VIS-XR) are shown in~\cref{fig-04:vis_irr,fig-05:vis_uv,fig-06:vis_xr}. For each example, the keypoint matches, the complete images (Fit to view option in the GUI), and a zoomed detail view in the synchronization mode are depicted showing good visual registration performance for all three multi-modal pairs. 

Secondly, we compare in \cref{tab-01} the success rates of ME and MAE of the registration for the C++ and PyTorch implementation of CraquelureNet using RANSAC for homography estimation or alternatively using USAC, GC-RANSAC, or MAGSAC++.  
The C++ implementation using RANSAC achieves comparable results as the PyTorch implementation (both $\tau_\text{reproj}=5$): For VIS-IRR and VIS-UV, all images are registered successfully using ME with an error threshold $\epsilon=5$ (PyTorch) and $\epsilon=7$ (C++). For VIS-XR, $12$ (C++) or $11$ (PyTorch) out of $13$ images were successfully registered using ME ($\epsilon=7$). 
The small deviations between the two models are due to slightly different implementations \eg of non-maximum suppression that were necessary for the conversion to C++.
Using the more recent USAC, GC-RANSAC, or MAGSAC++ is slightly less robust for VIS-IRR and VIS-UV, since not more than $12$ out of $13$ image pairs can be registered using ME ($\epsilon=7$) as one image pair fails completely. VIS-XR registration is the most difficult part of the three domain pairs due to the visually highly different appearance of the VIS and XR images. Here, we can observe similar performance of ME, with a slightly higher percentage of USAC, GC-RANSAC, and especially MAGSAC++ for $\epsilon=3$, but for $\epsilon=7$ those are on par with the PyTorch model and are slightly inferior to the C++ model using RANSAC.
Regarding the success rates of MAE, we observe a bit higher values for USAC, GC-RANSAC, and MAGSAC++ than for both RANSAC models for VIS-UV and VIS-XR ($\epsilon=\{8,10\}$), while for VIS-IRR they are the same.

In \cref{fig-07:eval_paintings_sota}, the multi-modal registration performance of our C++ CraquelureNet (RANSAC) is measured in comparison to SIFT~\cite{LoweDG2004}, SuperPoint~\cite{DeToneD2018}, and D2-Net~\cite{DusmanuM2019}. CraquelureNet achieves the highest success rates of ME and MAE for all multi-modal pairs. D2-Net is relatively close to CraquelureNet for VIS-IRR with the same SR of ME but with a lower SR of MAE. For VIS-IRR and VIS-UV, all learning-based methods are clearly better than SIFT. For the challenging VIS-XR domain, the advantage of our cross-modal keypoint detector and descriptor is most distinct, since for $\epsilon=7$, CraquelureNet still achieves high success rates of $92.3$\,\% for ME and $61.5$\,\% for MAE, whereas for $\epsilon=7$, D2-Net only successfully registers $61.5$\,\% for ME and $15.3$\,\% for MAE, SuperPoint only $23$\,\% for ME and none for MAE, and lastly, SIFT does not register any VIS-XR image pair successfully.
D2-Net was developed to find correspondences in difficult image conditions, such as day-to-night or depiction changes~\cite{DusmanuM2019}, hence it also is able to detect to some extend matching keypoint pairs in VIS-XR images. SuperPoint's focus is more on images with challenging viewpoints~\cite{DeToneD2018,DusmanuM2019}, thus the pretrained model results in a lower VIS-XR registration performance. In our prior work~\cite{SindelA2021}, we fine-tuned SuperPoint using image patches extracted from our manually aligned multi-modal paintings dataset, which did not result in an overall improvement. On the other hand, CraquelureNet is robust for the registration of all multi-modal pairs and does not require an intensive training procedure, as it is trained in efficient time only using very small image patches. 

The execution time of CraquelureNet (C++ and PyTorch) for the VIS-IRR registration (including loading of network and images, patch extraction of size $1024 \times 1024 \times 3$ pixels, network inference, homography estimation using RANSAC and image warping) was about 11\,s per image pair on the GPU and about 2\,min per image pair on the CPU. We used an Intel Xeon W-2125 CPU 4.00 GHz with 64 GB RAM and one NVIDIA Titan XP GPU for the measurements. 
The comparable execution times of the C++ and PyTorch implementation of CraquelureNet (the restructured one for a fair comparison) is not surprising, as PyTorch and OpenCV are wrappers around C++ functions.
The relatively fast inference time for the registration of high-resolution images makes CraquelureNet suitable to be integrated into the registration GUI and to be used by art technologists and art historians for their daily work.

\subsection{Registration of Historical Prints}
To evaluate the registration performance for prints, we have created a test dataset of in total $52$ images of historical prints with manually labeled control point pairs. The dataset is composed of $13$ different motifs of 16th century prints with each four exemplars that may show wear or production-related differences in the print. For each motif a reference image was selected and the other three copies will be registered to the reference, resulting in $39$ registration pairs. For each image pair, $10$ control point pairs were manually annotated.

We test the CraquelureNet, that was solely trained on paintings, for the 16th century prints. One qualitative example is shown in~\cref{fig-08:print}. 
CraquelureNet also finds a high number of good matches in this engraving pair, due to the multitude of tiny lines and branchings in the print as CraquelureNet is mainly focusing on these branching points.

\begin{figure}[t]
\centering
\begin{tikzpicture}
	\begin{customlegend}[
			legend entries={SIFT,D2-Net,SuperPoint,CraquelureNet (C++)},
			legend style={
				/tikz/every even column/.append	style={column sep=.75cm}},
			legend columns=-1,
			legend cell align=left]
		\addlegendimage{oorange,mark=square*}
		\addlegendimage{red,mark=*}		
		\addlegendimage{blue,mark=triangle*}
		\addlegendimage{cyan,mark=star}				
	\end{customlegend}
\end{tikzpicture}

\subcaptionbox{~\label{fig-09:prints_SR_ME}}{
\begin{tikzpicture}
\begin{axis} [
			scale = 0.6, 			
			xmajorgrids,
			ymajorgrids,
			scaled ticks=false,
			ylabel = {SR of ME [\%]},
			xlabel = Error threshold,
			xtick ={1,2,3,4,5,6},
			ymax   = 100,	
			ymin   = 0	
			]
	\addplot [oorange,mark=square*]
        table[x=Thresh, y=SIFT] 
        {tables/prints_success_rates_me.dat}; 	
	\addplot [red,mark=*]
        table[x=Thresh, y=D2-Net] 
        {tables/prints_success_rates_me.dat}; 
	\addplot [blue,mark=triangle*]
        table[x=Thresh, y=SuperPoint] 
        {tables/prints_success_rates_me.dat};
	\addplot [cyan,mark=star]
        table[x=Thresh, y=CraquelureNet] 
        {tables/prints_success_rates_me.dat};        
\end{axis}
\end{tikzpicture}
}
\hfill
\subcaptionbox{~\label{fig-09:prints_SR_MAE}}{
\begin{tikzpicture}
\begin{axis} [
			scale = 0.6, 			
			xmajorgrids,
			ymajorgrids,
			scaled ticks=false,
			ylabel = {SR of MAE [\%]},
			xlabel = Error threshold,
			xtick ={5,6,7,8,9,10},
			ymax   = 100,	
			ymin   = 0	
			]
	\addplot [oorange,mark=square*]
        table[x=Thresh, y=SIFT] 
        {tables/prints_success_rates_mae.dat}; 	
	\addplot [red,mark=*]
        table[x=Thresh, y=D2-Net] 
        {tables/prints_success_rates_mae.dat}; 
	\addplot [blue,mark=triangle*]
        table[x=Thresh, y=SuperPoint] 
        {tables/prints_success_rates_mae.dat};
	\addplot [cyan,mark=star]
        table[x=Thresh, y=CraquelureNet] 
        {tables/prints_success_rates_mae.dat};      
\end{axis}
\end{tikzpicture}
}
\caption{Quantitative comparison of success rates for registration of prints ($39$ image pairs). For all methods RANSAC with $\tau_\text{reproj}=5$ was used. None of the methods was fine-tuned for the print dataset.
In (\subref{fig-09:prints_SR_ME}) the success rate of mean Euclidean error (ME) for the error thresholds $\epsilon=\{1,2,...,6\}$ and in (\subref{fig-09:prints_SR_MAE}) the success rate of maximum Euclidean error (MAE) for the error thresholds $\epsilon=\{5,6,...,10\}$ is plotted.} 
\label{fig-09:prints_SR}
\end{figure}
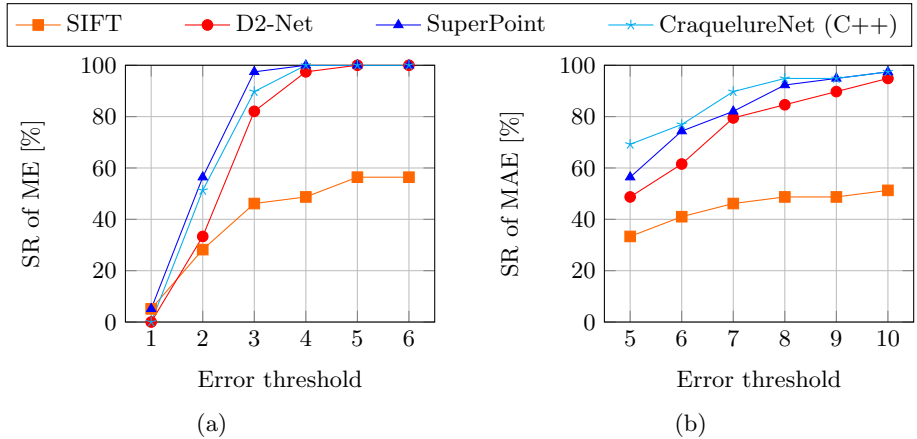

For the quantitative evaluation, we compute the success rates of ME and MAE for the registration of the print dataset using our CraquelureNet C++ model in comparison to using SIFT~\cite{LoweDG2004}, SuperPoint~\cite{DeToneD2018}, and D2-Net~\cite{DusmanuM2019}.
For the registration, the images are scaled to a fixed height of $2000$ pixel, as then the structures in the prints have a suitable size for the feature detectors. In the GUI of CraquelureNetApp this option is ``resize to custom height''. The results of the comparison are depicted in~\cref{fig-09:prints_SR}. The CNN-based methods obtain clearly superior results to SIFT for both metrics. Overall, CraquelureNet and SuperPoint show the best results, as both achieve a success rate of ME of $100$\,\% at error threshold $\epsilon=4$, where they are closely followed by D2-Net at $\epsilon=5$ and all three methods achieve a success rate of MAE close to $100$\,\% at $\epsilon=10$. For smaller error thresholds, CraquelureNet is slightly superior for MAE and SuperPoint for ME. As none of the methods was fine-tuned for the print dataset, this experiment shows the successful possibility of applying the models to this new dataset. 

\section{Conclusion}
We presented an interactive registration and visualization tool for multi-modal paintings and also applied it to historical prints. The registration is performed fully automatically using CraquelureNet. The user can choose the registration settings and can interact with the visualizations of the registration results. In the future, we could extend the application by including trained models of CraquelureNet on other datasets, such as the RetinaCraquelureNet~\cite{SindelA2022} for multi-modal retinal registration. A further possible extension would be to add a batch processing functionality to the GUI to register a folder of image pairs.

\subsubsection{Acknowledgements} 
Thanks to Daniel Hess, Oliver Mack, Daniel G\"orres, Wib\-ke Ottweiler, Germanisches Nationalmuseum (GNM), and Gunnar Heydenreich, Cranach Digital Archive (CDA), and Thomas Klinke, \mbox{TH K\"oln}, and Amalie H\"ansch, FAU Erlangen-N\"urnberg for providing image data, and to Leibniz Society for funding the research project ``Critical Catalogue of Luther portraits (1519 - 1530)'' with grant agreement No. SAW-2018-GNM-3-KKLB, to the European Union’s Horizon 2020 research and innovation programme within the Odeuropa project under grant agreement No. 101004469 for funding this publication, and to NVIDIA for their GPU hardware donation. 

%
%
%
 \bibliographystyle{splncs04}
 \bibliography{refs}

\end{document}